\documentclass[letterpaper, 10 pt, journal,twoside]{IEEEtran}
\IEEEoverridecommandlockouts
\let\labelindent\relax
\let\labelindent\relax
\usepackage{graphicx} \graphicspath{ {figures/} }
\usepackage{amsmath,amssymb,mathabx,amsbsy,mathtools,etoolbox}
\usepackage[utf8]{inputenc} 
\usepackage[T1]{fontenc}    

\usepackage{algorithmic}
\usepackage[ruled]{algorithm2e}
\usepackage{acronym}
\usepackage{enumitem}
\usepackage[breaklinks,colorlinks,linkcolor=black]{hyperref}
\usepackage{balance}
\usepackage{xspace}
\usepackage{setspace}
\usepackage[skip=3pt,font=small]{subcaption}
\usepackage[skip=3pt,font=small]{caption}
\usepackage[dvipsnames,svgnames,x11names]{xcolor}
\usepackage[capitalise,nameinlink]{cleveref}
\usepackage{booktabs,tabularx,colortbl,multirow,multicol,array,makecell}
\usepackage{pifont}
\usepackage{cite}
\usepackage{tikz}
\usepackage{nicematrix}
\usepackage{dblfloatfix}

\makeatletter
\DeclareRobustCommand\onedot{\futurelet\@let@token\@onedot}
\def\@onedot{\ifx\@let@token.\else.\null\fi\xspace}

\def\etal{\emph{et al}\onedot}
\makeatother

\makeatletter
\def\BState{\State\hskip-\ALG@thistlm}
\makeatother

\makeatletter
\renewcommand{\paragraph}{%
  \@startsection{paragraph}{4}%
  {\z@}{0ex \@plus 0ex \@minus 0ex}{-1em}%
  {\hskip\parindent\normalfont\normalsize\bfseries}%
}
\makeatother

\crefname{algorithm}{Alg.}{Algs.}
\Crefname{algocf}{Algorithm}{Algorithms}
\crefname{section}{Sec.}{Secs.}
\Crefname{section}{Section}{Sections}
\crefname{table}{Tab.}{Tabs.}
\Crefname{table}{Table}{Tables}
\crefname{figure}{Fig.}{Fig.}
\Crefname{figure}{Figure}{Figure}

\definecolor{gblue}{HTML}{4285F4}
\definecolor{gred}{HTML}{DB4437}
\definecolor{ggreen}{HTML}{0F9D58}

\definecolor{mygray}{gray}{.92}

\acrodef{dof}[DoF]{Degree of Freedom}
\acrodef{ik}[IK]{Inverse Kinematics}
\acrodef{mpc}[MPC]{model predictive control}
\acrodef{wbc}[WBC]{whole-body control}
\acrodef{ros}[ROS]{Robot Operating System}
\acrodef{slam}[SLAM]{Simultaneous Localization and Mapping}
\acrodef{com}[CoM]{center of mass}
\acrodef{cam}[CAM]{centroidal angular momentum}
\acrodef{cmm}[CMM]{centroidal momentum matrix} 
\acrodef{cdm}[CDM]{centroidal dynamics model}
\acrodef{cop}[CoP]{Center of Pressure}
\acrodef{srbm}[SRBM]{single rigid body model}
\acrodef{lipm}[LIPM]{linear inverted pendulum model}
\acrodef{slipm}[SLIPM]{spring-loaded inverted pendulum model}
\acrodef{ccrbi}[CCRBI]{centroidal composite rigid body inertia}

\title{CDM-MPC: An Integrated Dynamic Planning and Control Framework for Bipedal Robots Jumping}

\author{Zhicheng He, Jiayang Wu, Jingwen Zhang, Shibowen Zhang, Yapeng Shi, \\Hangxin Liu, Lining Sun, 
Yao Su,~\IEEEmembership{Member,~IEEE}, and Xiaokun Leng 

\thanks{Manuscript received February 4, 2024; revised April 21, 2024; accepted May 16, 2024. Date of publication xx May 2024; date of current version xx May 2024. This paper was recommended for publication by Editor Abderrahmane Kheddar upon evaluation of the Associate Editor and Reviewers' comments. This work was supported in part by the National Natural Science Foundation of China (No.52305072), Natural Science Foundation of Hebei Province of China (No.E2022203095), and Shenzhen Special Fund for Future Industrial Development (No.KJZD20230923114222045). (\textit{Corresponding authors: Yao Su and Xiaokun Leng.})} 
\thanks{Zhicheng He, Jiayang Wu, Yapeng Shi and Xiaokun Leng are with department of Computer Science, Harbin Institute of Technology, Harbin 150001, China (e-mails: hezhicheng@hit.edu.cn; 2021113679@stu.hit.edu.cn; shi.yapeng@hit.edu.cn; lengxiaokun@hit.edu.cn).}
\thanks{Jingwen Zhang, Shibowen Zhang, Hangxin Liu and Yao Su are with State Key Laboratory of General Artificial Intelligence, Beijing Institute for General Artificial Intelligence (BIGAI), Beijing 100080, China (e-mails: zhangjingwen@bigai.ai; zhangshibowen@bigai.ai; liuhx@bigai.ai; suyao@bigai.ai).}
\thanks{Lining Sun is with School of Mechatronics Engineering, Harbin Institute of Technology, Harbin 150080, China, and also with Jiangsu Provincial Key Laboratory of Advanced Robotics, School of Mechanical and Electric Engineering, Soochow University, Suzhou 215000, China (e-mail: lnsun@hit.edu.cn).}
\thanks{This letter has supplementary downloadable material available at https://doi.org/10.1109/LRA.2024.xxxxxxx, provided by the authors.}
\thanks{Digital Object Identifier (DOI): see top of this page.}}

\begin{document}

\maketitle

\begin{abstract}
Performing acrobatic maneuvers like dynamic jumping in bipedal robots presents significant challenges in terms of actuation, motion planning, and control. Traditional approaches to these tasks often simplify dynamics to enhance computational efficiency, potentially overlooking critical factors such as the control of centroidal angular momentum (CAM) and the variability of centroidal composite rigid body inertia (CCRBI). This paper introduces a novel integrated dynamic planning and control framework, termed centroidal dynamics model-based model predictive control (CDM-MPC), designed for robust jumping control that fully considers centroidal momentum and non-constant CCRBI. The framework comprises an optimization-based kinodynamic motion planner and an MPC controller for real-time trajectory tracking and replanning. Additionally, a centroidal momentum-based inverse kinematics (IK) solver and a landing heuristic controller are developed to ensure stability during high-impact landings. The efficacy of the CDM-MPC framework is validated through extensive testing on the full-sized humanoid robot KUAVO in both simulations and experiments.
\end{abstract}


\setstretch{0.96}
\section{Introduction}\label{sec:intro}
\IEEEPARstart{A}{chieving} acrobatic motions, a significant challenge in bipedal robotics, requires not only powerful robot actuators~\cite{chignoli2021humanoid,zhu2019design} but also sophisticated motion planning and control algorithms~\cite{dai2014whole,herzog2016structured}. Unlike the control of walking or running\textemdash{}where the \textbf{\ac{cam}} is typically overlooked to simplify the highly nonlinear multi-body dynamics using models like \ac{lipm}, \ac{slipm}, or \ac{srbm} for computational efficiency, \ac{cam} plays a key role in the jumping control of bipedal robots. This introduces unique challenges and necessitates additional considerations in the control strategy to accurately manage \ac{cam} throughout the entire process~\cite{saccon2017centroidal, zhang2023design}. 

\begin{figure}[t!]
    \centering
    \includegraphics[width=\linewidth,trim=1cm 1cm 1cm 1cm, clip]{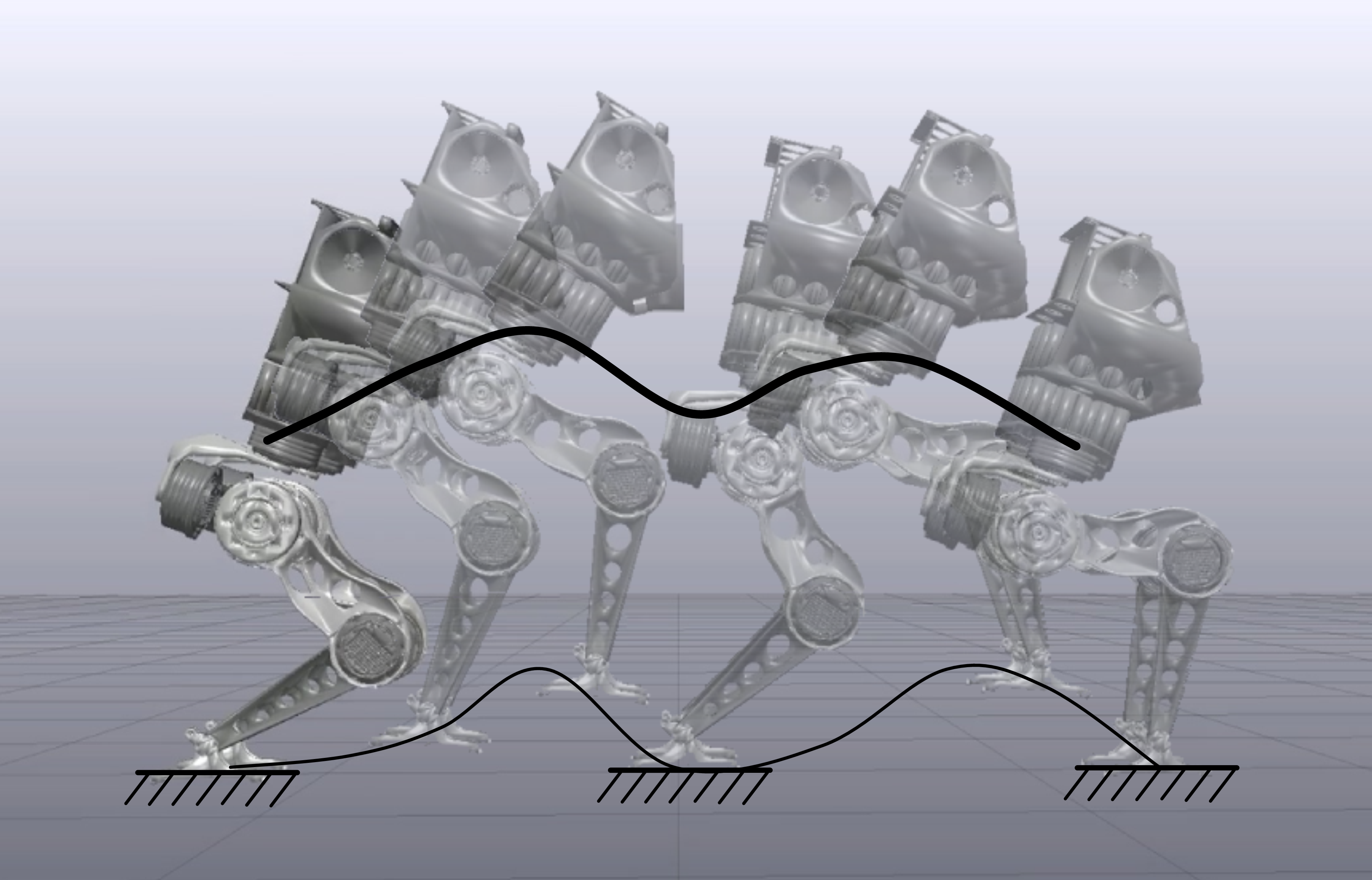}
    \caption{\textbf{The proposed integrated dynamic planning and control framework endows bipedal robots capable of continuously forward jumping.} The trajectories of the foot and the torso links are plotted in thin and bold black lines, respectively.}
    \label{fig:motiv}
\end{figure}

\begin{figure*}[b!]
    \centering
    \includegraphics[width=0.9\linewidth,trim=0cm 0cm 3cm 0cm, clip]{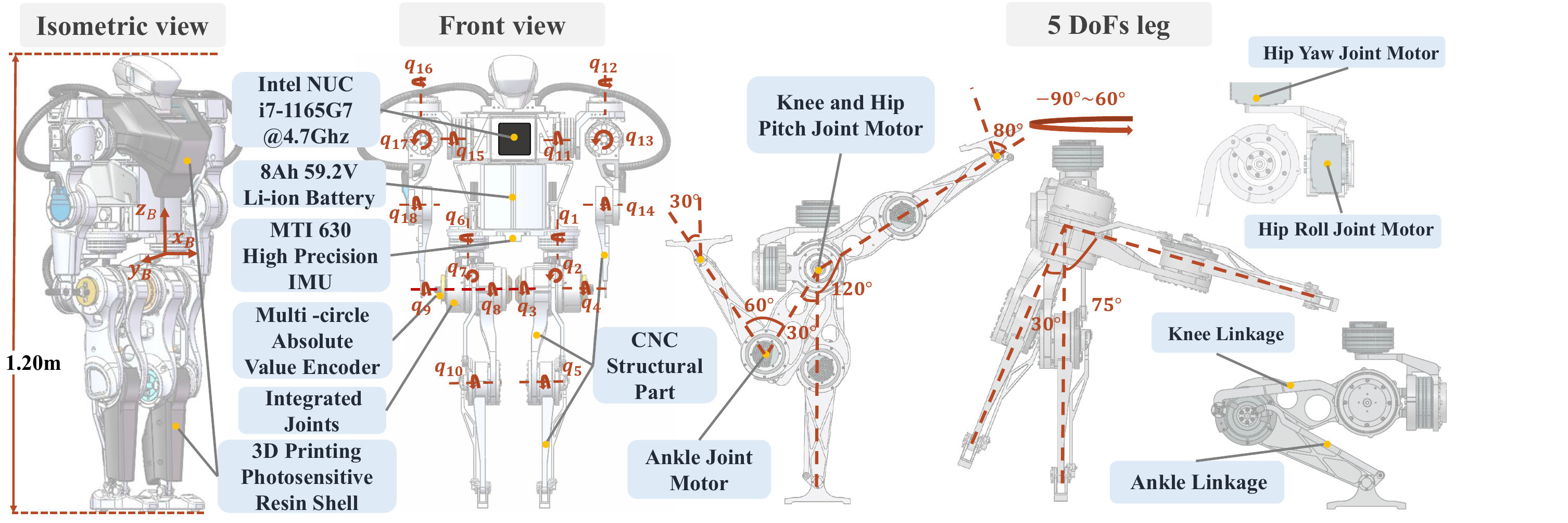}
    \caption{\textbf{Hardware design and configuration of the bipedal humanoid robot KUAVO.} Each leg contains 5 \acp{dof}: 3 \acp{dof} for the hip joint, 1 \ac{dof} for the knee joint and 1 \ac{dof} for the ankle joint.}
    \label{fig:hardware}
\end{figure*}

The dynamic jumping process of bipedal robots, depicted in \cref{fig:motiv}, consists of three distinct phases: the \textbf{launching phase}, the \textbf{flight phase}, and the \textbf{landing phase}. Each phase poses unique challenges in dynamics planning and control. Specifically, (i) the launching phase requires an accurate kinodynamic motion planner to generate trajectories for the \ac{com} and \ac{cam} within the hardware capabilities~\cite{wensing2024optimization}; (ii) the flight phase necessitates a fast trajectory replanning scheme to address motor execution errors and unknown disturbances, alongside an advanced \ac{wbc}-based \ac{cam} controller for accurate trajectory tracking~\cite{qi2023vertical,zhang2023design}; (iii) the landing phase demands a robust landing controller to preserve platform stability under substantial impact forces~\cite{zhang2023design,chignoli2021humanoid}.

Researchers have developed various integrated motion planning and control frameworks~\cite{zhang2023design,qi2023vertical,li2023robust,mesesan2023unified,saloutos2023design} to address this challenge. Specifically, Zhang \etal employed \ac{cam} for planning the entire jumping motion with a fixed foothold location and 
a heuristic landing planner to compensate for disturbances or tracking errors~\cite{zhang2023design}. Qi \etal designed a \ac{cam} controller based on \ac{lipm} for real-time planning of whole-body dynamics with predefined CoP~\cite{qi2023vertical}. Li \etal utilized a reinforcement learning (RL)-based approach to establish a robot-specific jumping control policy~\cite{li2023robust}.  Saloutos \etal and Mesesan \etal generated jumping trajectories through trajectory optimization leveraging full-order dynamics or a 3-D divergent component of motion (3D-DCM) framework, respectively~\cite{saloutos2023design,mesesan2023unified}. Despite their theoretical contributions, these frameworks have not achieved the continuous jumping of a humanoid robot, which requires more robustness than a single vertical jumping.

In this study, we introduce a novel integrated dynamic planning and control framework\textemdash{}centroidal dynamics model-based model predictive control (CDM-MPC), tailored for bipedal robot jumping. This framework thoroughly accounts for the full centroidal momentum while considering non-constant \ac{ccrbi} to achieve precise body pose control. Initially, we formulate an optimization-based kinodynamic motion planner that concurrently optimizes the \ac{cam} trajectory, touchdown point, and contact force trajectories. Subsequently, we design an innovative \ac{mpc} controller for precise trajectory tracking and fast online replanning. Further, we develop a centroidal momentum-based \ac{ik} solver to achieve accurate centroidal momentum tracking and a landing heuristic controller to ensure robust stabilization during high-impact landings. The efficacy of the proposed framework is corroborated through rigorous testing on our full-scale humanoid robot platform, KUAVO, within both realistic simulation environments and real-world experimental settings.

\subsection{Related Work}\label{sec:related}

\textbf{\ac{mpc}-based motion planning/control frameworks} have gained significant attention in recent years due to their capability to account for kinematics, dynamics, physical, and contact constraints through a unified optimization formulation. These frameworks have demonstrated success across various robotic configurations, including wheeled robots~\cite{zhou2024aspire,gao2024probabilistic}, UAVs~\cite{su2024real}, bipedal robots~\cite{chignoli2021humanoid, kuindersma2016optimization, luo2019robust}, quadruped robots~\cite{bledt2018cheetah, kim2019highly,mastalli2022agile,zhou2022momentum}, and hexapod robots~\cite{wang2020walkingbot, lin2019optimization}. In the realm of bipedal robots, Pratt \etal first proposed the capture point method to determine the ideal footstep locations post-disturbance within an \ac{mpc} framework~\cite{pratt2006capture}, albeit lacking consideration of kinematic and dynamic constraints. Scianca \etal enhanced this approach by incorporating Zero Moment Point (ZMP), kinematics, and stability constraints for improved gait generation capability~\cite{scianca2020mpc}, while Brasseur \etal extends the method to a 3D walking scenario with online gait generation. Concurrently optimization of footstep locations and step time was achieved by a nonlinear \ac{mpc} formulation~\cite{kryczka2015online}. These methodologies excelled in low-speed walking scenarios by utilizing simplified models, such as \ac{lipm} and \ac{slipm} for computational efficiency, which is less suited for dynamic motions where the \ac{cam} of the robot can not be neglected.

To enhance the generation of dynamic motions for legged robots across various speeds and magnitudes, the \textbf{\ac{cdm}} was integrated into trajectory optimization formulations~\cite{budhiraja2019dynamics,ponton2021efficient,kwon2020fast}. Specifically, Budhiraja \etal introduced a formulation to reconcile the \ac{cdm} with the complete dynamics model~\cite{budhiraja2019dynamics}; Ponton \etal employed the \ac{cdm} for efficient walking pattern generation~\cite{ponton2021efficient}; Kwon \etal utilized the \ac{cdm} for footstep planning, integrating a momentum-mapped \ac{ik} solver to design whole-body motion~\cite{kwon2020fast}. Moreover, the \ac{cdm}-based real-time \ac{mpc} frameworks have been successfully implemented on agile maneuver control of quadruped robots, considering the constant \ac{ccrbi} of the robot~\cite{grandia2023perceptive,meduri2023biconmp}. In this study, we concentrate on the jumping control of biped robots while accounting for non-constant \ac{ccrbi}. We first derive the relationship between \ac{ccrbi} and robot leg length as a constraint, then incorporate it into a real-time \ac{mpc} framework to accurately regulate the body posture of a biped robot during the jumping process.   

\subsection{Overview}
We organize the remainder of the paper as follows. \cref{sec:dynamics} introduces the kinematics and dynamics model of the system. \cref{sec:planning} describes the dynamical planning and control framework. \cref{sec:verification} presents the simulation and experiment results with comprehensive evaluations. Finally, we conclude the paper in \cref{sec:conclusion}.

\section{Platform Model}
\label{sec:dynamics}
\subsection{Hardware Introduction}
\label{sec:hardware_introduction}
The hardware platform depicted in \cref{fig:hardware} is developed to evaluate the performance of the proposed CDM-MPC framework. The KUAVO bipedal robot platform stands $1.2~m$ in height and weighs $34.5~kg$. It incorporates 18 motors: each leg is equipped with 5 \acp{dof} and each arm with 4 \acp{dof}. A key design innovation implemented to optimize the robot's performance involves the strategic placement of the ankle joint motor at the knee and the knee joint motor at the hip. These modifications, coupled with linkage mechanisms, are intended to reduce leg inertia and improve the robot's maneuverability. The robot operates in a Linux system with an augmented real-time kernel patch to prompt the reading of sensor inputs, the whole system runs on an onboard Intel NUC computer (Intel Core i7-1165G7@4.7GHz). Notably, all joint actuators are configured to operate in the torque control mode, ensuring precise and efficient control. For a comprehensive overview of the hardware platform's specifications, please refer to \cref{tab:hardware}.

\begin{table}[t!]
    \small
    \centering
    \caption{\textbf{Main Physical Parameters of KUAVO Robot}}
    \label{tab:hardware}
    \resizebox{0.8\linewidth}{!}{%
        \begin{tabular}{c  c  c  c  c}
            \toprule
            \multicolumn{5}{c}{\textbf{Dimension Parameters}} \\
            \midrule
            Total mass & Pelvis width & Thigh length & Calf length &   Foot length \\
        
             34.5 [kg] & 0.22 [m] & 0.23 [m] & 0.26 [m] & 0.15 [m] \\
            \toprule
            \multicolumn{5}{c}{\textbf{Motion Range \& Joint Peak Torque}}\\
            \midrule
            Hip Yaw & Hip Roll & Hip Pitch & Knee Pitch &  Ankle Pitch\\
            $-90^{\degree} \sim 60^{\degree}$ & $-30^{\degree} \sim 75^{\degree}$ &  $-30^{\degree} \sim 120^{\degree}$ & $-120^{\degree} \sim 10^{\degree}$ & $-30^{\degree} \sim 80^{\degree}$\\
            48 [Nm] &110 [Nm] &110 [Nm] &110 [Nm] &48 [Nm] \\
            \bottomrule
        \end{tabular}}%
\end{table}

\subsection{Dynamics}
\subsubsection{Floating Base Dynamics} 
The comprehensive dynamics of a biped robot can be described as: 
\begin{equation}
    \pmb{M}(\pmb{q})\pmb{\ddot{q}} + \pmb{C}( {\pmb{q},\pmb{\dot q}}) = \pmb{\Lambda}\,\pmb{\tau}  + {\pmb{J}^\mathsf{T}}(\pmb{q})\pmb{f}_{\text{ext}},
\label{eq:equations_of_motion}
\end{equation}
where $\pmb{M}(\pmb{q})$ is the inertia matrix, $\pmb{C}\left( {\pmb{q},\pmb{\dot q}} \right)$ is the vector of Coriolis, centrifugal and gravitational forces, $\pmb{\tau}$ is the joint torque, $\pmb{\Lambda}$ is the input mapping matrix, $\pmb{J}(\pmb{q})$ is the Jacobian matrix and $\pmb{f}_{\text{ext}}$ is the external force vector~\cite{luo2021modeling}. However, when utilizing this highly nonlinear multi-body dynamics model for controller design, computational efficiency becomes a significant concern. Even with the advanced nonlinear optimizers~\cite{gill2005snopt}, real-time control applications remain problematic. 

\subsubsection{Centroidal Dynamics} 
Consequently,  the \ac{cdm} proposed by Orin \etal~\cite{Orin2013Centroidal} is employed in this work, offering a holistic approach to capturing whole body momentum while maintaining simplicity.  It can be formally written as:
\begin{equation}
    \pmb{H}=
    \begin{bmatrix}
       \pmb{h} \\ m\pmb{\dot{r}}
    \end{bmatrix}=
    \pmb{A}(\pmb{q}) \pmb{\dot{q}},
    \label{eq:cmcal}
\end{equation}
where the centroidal momentum vector $\pmb{H}\in\mathbb{R}^{6\times1}$ consists of the angular centroidal momentum $\pmb{h}\in\mathbb{R}^{3\times1}$ and the linear centroidal momentum $m\pmb{\dot{r}}$ of the robot. Here, $m$ denotes the total mass and $\pmb{r}\in\mathbb{R}^{3\times1}$ represents the \ac{com} position, $\pmb{A}\in\mathbb{R}^{6{\times}16}$ is the \ac{cmm}, $\pmb{q}\in\mathbb{R}^{16{\times}1}$ is the generalized position vector. Notably, for model simplicity the dynamics and control of the arms are not considered in work, the whole upper body of the robot is treated as one single body. Therefore, $\pmb{q}$ includes the position and attitude of the base (6-\acp{dof}), and the 10-\acp{dof} of the legs.

\begin{figure}[t!]
  \centering
  \includegraphics[width=0.45\linewidth, trim=0cm 0.5cm 0cm 0.3cm, clip]{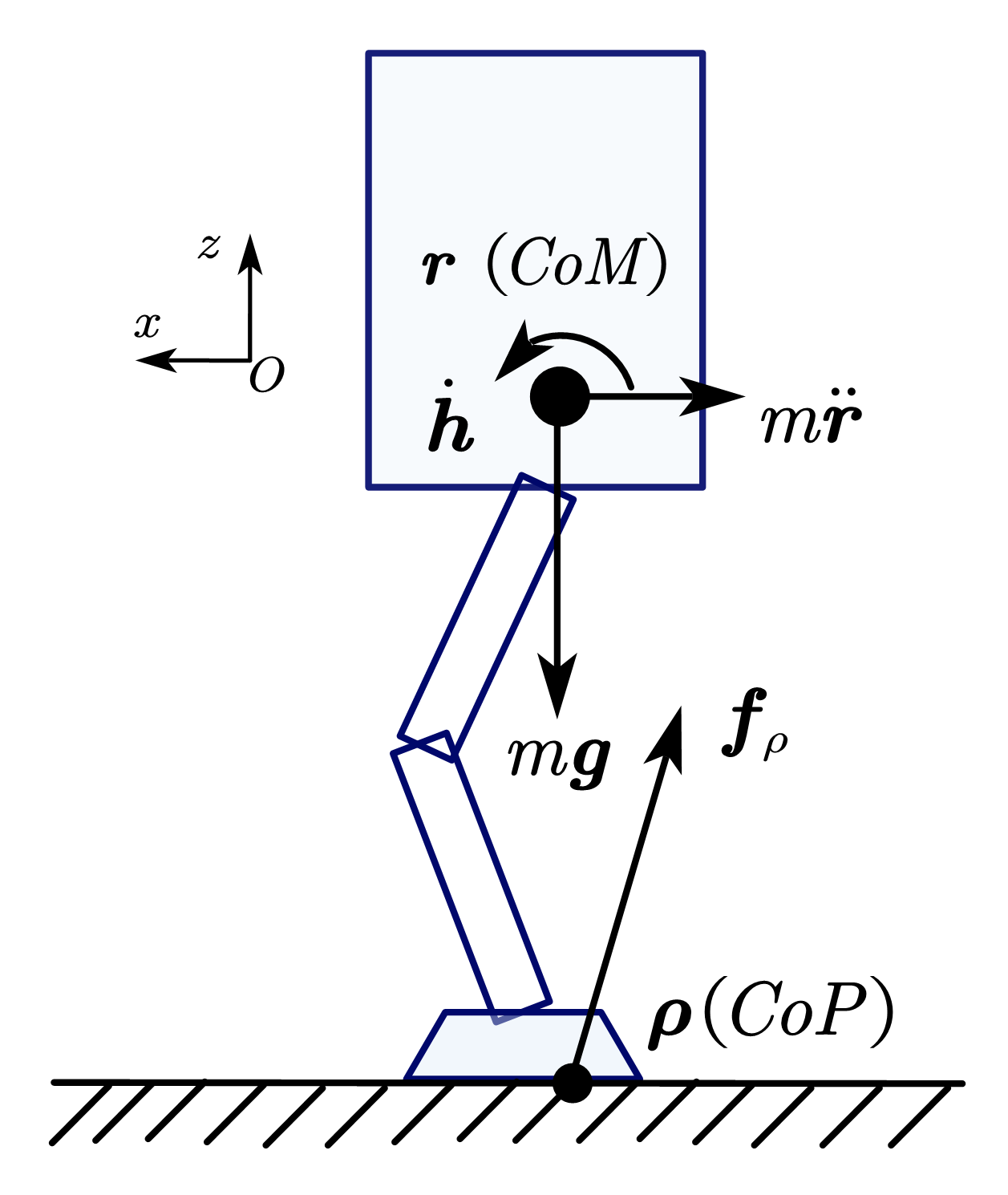}
  \caption{\textbf{External forces analysis.} In the $xOz$ plane, the robot is subjected to gravitational force $m\pmb{g}$ and ground reaction force $\pmb{f}_{\rho}$.}
  \label{fig:CentroidalDynamics}
\end{figure}

According to Newton's laws of motion, the rate of change of angular and linear momentum at the \ac{com}, $\dot{\pmb{h}}$ and $m{\pmb{\ddot{r}}}$, is equivalent to the resultant effect of all external forces. As illustrated in \cref{fig:CentroidalDynamics}, we can have:
\begin{equation}
\begin{aligned}
  \pmb{\dot{h}} &=(\pmb{\rho}-\pmb{r}) \times \pmb{f}_{\rho},\\
  m{\pmb{\ddot{r}}}&=m\pmb{g}+\pmb{f}_{\rho},\\
\end{aligned}
\label{eq:momentrate}
\end{equation}
where $m\pmb{g}$ and $\pmb{f}_{\rho}\in\mathbb{R}^{3\times1}$ are the gravitational force and ground reaction force of the robot, respectively. $\pmb{\rho}\in\mathbb{R}^{3\times1}$ is the \ac{cop} location.  

\begin{figure*}[t!]
    \centering
    \includegraphics[width=\linewidth,trim=0cm 2.7cm 0cm 1.0cm, clip]{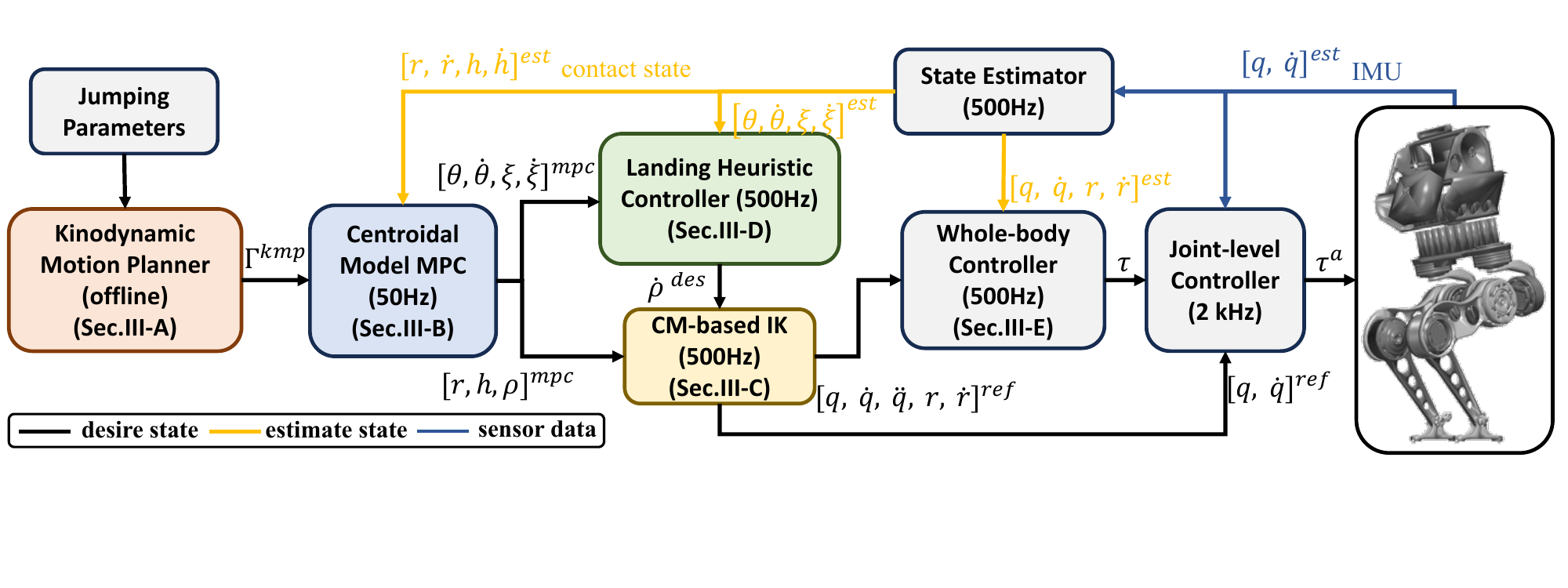}
    \caption{\textbf{The CDM-MPC dynamic planning and control framework.} (i) The CDM-based kinodynamic motion planner produces the centroidal momentum reference trajectory. (ii) The real-time MPC controller provides accurate trajectory tracking and fast replanning under disturbances. (iii) The centroidal moment-based IK solves whole-body trajectory without simplifying leg dynamics. (iv) The landing heuristic controller guarantees robust landing stabilization.}
    \label{fig:diagram}
\end{figure*}

\section{Dynamic Planning \& Control Framework} \label{sec:planning}
Building upon the \ac{cdm}, we develop the CDM-MPC framework to address the complexities encountered during bipedal robot jumping in the launching, flight, and landing phases, as introduced in \cref{sec:intro}. Our framework consists of four primary components: (i) an optimization-based kinodynamic motion planner to produce the \ac{cdm} trajectory, contact force, and contact position; (ii) a real-time MPC controller with \ac{cdm} for trajectory tracking and fast replanning; (iii) an \ac{ik} solver based on centroidal momentum that calculates whole-body trajectories; (iv) a landing heuristic controller for robust stabilization. The overall framework is summarized in \cref{fig:diagram}.

\subsection{Kinodynamic Motion Planning} \label{subsec:trajplanning}
Assuming each foot has two contact points—the heel and the toe, with the contact force at each point constrained by a linearized friction cone comprising four edges, we formulate the kinodynamic motion planning as a trajectory optimization problem and define the set of decision variables as:
\begin{equation}
\resizebox{0.91\linewidth}{!}{
$\begin{aligned}
&\pmb{\Gamma}^{\text{kmp}}=\{\pmb{q}_{\left[k\right]},\pmb{\dot{q}}_{\left[k\right]},\pmb{r}_{\left[k\right]},\pmb{\dot{r}}_{\left[k\right]}, \pmb{\ddot{r}}_{\left[k\right]},\pmb{h}_{\left[k\right]}, \pmb{\dot{h}}_{\left[k\right]}, \pmb{f}_{\rho \left[k\right]}^i, \pmb{\rho}_{\left[k\right]}^i,\beta_{\left[k\right]}^{ij}, 
\\& \forall i=1,\cdots,N_c, j=1,\cdots,N_d, k=1, \cdots,N \} \in\mathbb{R}^{87\times N},
\end{aligned}$}
\end{equation}
where the subscript $\left[k\right]$ represents the $k^{th}$ time interval, $N$ is the total number of timesteps, $\pmb{\rho}^{i}$ and $\pmb{f}_{\rho}^{i}$ denote the $i^{th}$ contact location and corresponding contact force, $N_c=4$ is the number of contact points, $\beta^{ij}$ represents the portion of the friction along each edge of the linearized friction pyramid, $N_d=16$ is the total number of these approximated edges.

The kinematic constraints include~\cite{dai2014whole}
\begin{subequations}
\begin{align}
    \pmb{H}_{\left[k\right]}&=\pmb{A}(\pmb{q}_{\left[k\right]}) \pmb{\dot{q}}_{\left[k\right]},\label{eq:planning_con3}\\
    \pmb{r}_{\left[k\right]}&=\pmb{\mathcal{F}} _r(\pmb{q}_{\left[k\right]}),\label{eq:planning_con4}\\
\pmb{\rho}_{\left[k\right]}^{i}&=\pmb{\mathcal{F}}_{\rho}^{i}(\pmb{q}_{\left[k\right]}), \quad\quad\quad i=1,\cdots,N_c, \label{eq:planning_con5}\\
    \left\|\pmb{\rho}_{\left[k\right]}^{i}-\pmb{\rho}_{\left[k-1\right]}^{i} \right\| &\leq \pmb{D}\left( 1-\pmb{S}^i_{\left[k\right]} \right), \quad i=1,\cdots,N_c,  \label{eq:planning_con7}\\
    \left\|\pmb{r}_{\left[k\right]}-\pmb{\rho}_{\left[k\right]}^{i} \right\| &\geq{\xi}_{\text{min}}, \quad\quad\quad\quad\quad i=1,\cdots,N_c, \label{eq:planning_con8}
\end{align}
\label{eq:kine_cons}
\end{subequations}
where $\pmb{\mathcal{F}}_r(\cdot)$ and $\pmb{\mathcal{F}}_\rho^i(\cdot)$ are forward kinematics functions to compute $\pmb{r}$ and $\pmb{\rho}^i$, $\pmb{S}^i$ is a binary variable representing the predefined contact sequence, $\pmb{D}$ is the maximum allowable distance between consecutive timesteps, $\xi_{\text{min}}$ is the minimal distance between $\pmb{r}$ and $\pmb{\rho}^i$. Of note, \cref{eq:planning_con7} limits the stepsize of each contact point while avoiding the movement of the contact point which is in contact. \cref{eq:planning_con8} is an approximate collision avoidance constraint to keep the minimal safety distance without adding computational complexity.

The dynamics constraints include
\begin{subequations}
\begin{align}
   \pmb{\dot{h}}_{\left[k\right]}&=\sum_{i=1}^{N_c}
	(\pmb{\rho}_{\left[k\right]}^{i}-\pmb{r}_{\left[k\right]}) \times \pmb{f}_{\rho [k]}^{i},\label{eq:planning_con1}\\
    \pmb{\ddot{r}}_{\left[k\right]}&=\frac{1}{m}{{\sum_{i=1}^{N_c}{\pmb{f}_{\rho \left[k\right]}^{i}}}},\label{eq:planning_con2}\\
    \pmb{f}_{\rho \left[k\right]}^{i}&=\sum_{j=1}^{N_d}{\beta_{\left[k\right]}^{ij}\pmb{v}_{\left[k\right]}^{ij}}, i=1,\cdots,N_c, j=1,\cdots,N_d,\label{eq:planning_confircton1}\\
    \beta_{\left[k\right]}^{ij}&\geq 0, \quad\quad\quad\quad i=1,\cdots,N_c, j=1,\cdots,N_d, \label{eq:planning_confircton2}\\
    \left\| \pmb{f}_{\rho \left[k\right]}^{i} \right\| &\leq \,f_{\text{max}}\pmb{S}^i_{\left[k\right]}, \quad i=1,\cdots,N_c,\label{eq:planning_con6} 
\end{align}
\label{eq:dyn_cons}
\end{subequations}
where the unit vector $\pmb{v}^{ij}$ denotes the $j^{th}$ edge of the friction pyramid at the $i^{th}$ contact point, $f_{\text{max}}$ is the maximum contact force for each contact point. The complete formulation is as follows:  
\begin{equation}
\resizebox{0.91\linewidth}{!}{
$\begin{aligned}
   &\min_{\pmb{\Gamma}^{\text{kmp}}} \sum_{k=1}^N{\left(\sum_{i=1}^{N_c}{\left\|\pmb{f}_{\rho \left[k\right]}^i\right\|^2}+\left\| \pmb{\ddot{r}}_{\left[k\right]}\right\|^2+\left\|\pmb{\dot{h}}_{\left[k\right]}\right\|^2+\left\|  \pmb{\dot{q}}_{\left[k\right]} \right\|^2 +\left\|\pmb{q}_{\left[k\right]}- \pmb{q}^* \right\|^2\right)}\\
   &\textit{s.t.} \quad\quad\quad\quad\quad\quad\quad\quad\quad\quad    \text{\cref{eq:dyn_cons,eq:kine_cons}}
   \end{aligned}$}
    \label{eq:trajopt}
\end{equation}
where the object function includes (i) a smoothness cost  $\left\| \pmb{\ddot{r}}_{\left[k\right]}\right\|^2+\left\|\pmb{\dot{h}}_{\left[k\right]}\right\|^2$ which penalizes for the rate of change in centroidal momentum; (ii) a contact force cost $\sum_{i=1}^{N_c}{\left\|\pmb{f}_{\rho \left[k\right]}^i\right\|^2}$ that minimizes the contact force; and (iii) a tracking cost  $\left\|\pmb{q}_{\left[k\right]}- \pmb{q}^* \right\|^2+\left\| \pmb{\dot{q}}_{\left[k\right]} \right\|^2 $ to regularize the solution around the nominal posture $\pmb{q}^*$.

\subsection{Real-time MPC Controller} 
To achieve real-time replanning after the launching phase, a \ac{mpc} controller is designed by simplifying the trajectory planning schema \cref{eq:trajopt} for computing efficiency improvement. Diverging from other \ac{mpc} formulations~\cite{saloutos2023design,budhiraja2019dynamics,ponton2021efficient,kwon2020fast}, our \ac{mpc} controller design accounts for non-constant leg inertia throughout the flight phase and accurately manipulates the body pose through actuated \ac{ccrbi} regulation~\cite{lee2007reaction}.

\subsubsection{Centroidal Inertia Decomposition}
In centroidal dynamics, the average spatial velocity is defined as~\cite{Orin2013Centroidal}: 
\begin{equation}
    \pmb{h}
    =
    \prescript{\bar{\omega}}{}{\pmb{I}}_{h}(\pmb{q})\,\pmb{\bar{\omega}}, 
\label{eq:avg_omega}
\end{equation}
where $\pmb{\bar{\omega}}$ is the average angular velocity of \ac{com}, $\prescript{\bar{\omega}}{}{\pmb{I}}_{h}(\pmb{q})\in\mathbb{R}^{3\times3}$ is the rotational part of the \ac{ccrbi}, defined as the sum of all the link inertia projected to the \ac{com}, and can be divided into the floating-base and actuated parts:
\begin{equation}
    \resizebox{0.91\linewidth}{!}{
    $\begin{aligned}
\prescript{\bar{\omega}}{}{\pmb{I}}_{h}(\pmb{q})&=\pmb{X}_{h}(\pmb{q})^\mathsf{T}\pmb{I}\pmb{X}_{h}(\pmb{q})\\
        &=
    \pmb{X}_{h}^\mathsf{T}
    \begin{bmatrix}
        \pmb{I}_1&		&		&		\\
        &		0&		&		\\
        &		&		\ddots&		\\
        &		&		&		0\\
    \end{bmatrix}\pmb{X}_{h}
    +
    \pmb{X}_{h}^\mathsf{T}
    \begin{bmatrix}
        0&		&		&		\\
        &		\pmb{I}_2&		&		\\
        &		&		\ddots&		\\
        &		&		&		\pmb{I}_{N_L}\\
    \end{bmatrix} \pmb{X}_{h}\\
    &=\prescript{\bar{\omega}}{}{\pmb{I}}_{h}^f+\prescript{\bar{\omega}}{}{\pmb{I}}_{h}^a(\pmb{q}),
    \end{aligned}$}
    \label{eq:inertia_decomposition}
    \end{equation}
where $\pmb{X}_{h}$ projects motion vectors from centroidal coordinate frame to link coordinate frames, $\pmb{I}_i$ is the spatial inertia for link $i$ with $i=1$ refers to the body link, $N_L$ is the number of links. Within \ac{ccrbi}, $\prescript{\bar{\omega}}{}{\pmb{I}}_{h}^f$ is unactuated and invariant whereas $\prescript{\bar{\omega}}{}{\pmb{I}}_{h}^a(\pmb{q})$ changes according to the leg configuration. 

Combining \cref{eq:inertia_decomposition} with \cref{eq:avg_omega}, we obtain:
\begin{equation}
\label{eq:ccrbi_calculation}
\pmb{h}=\prescript{\bar{\omega}}{}{\pmb{I}}_{h}(\pmb{q})\,\pmb{\bar{\omega}}=(\prescript{\bar{\omega}}{}{\pmb{I}}_{h}^{f}+\prescript{\bar{\omega}}{}{\pmb{I}}_{h}^{a}(\pmb{q}))\,\pmb{\bar{\omega}},
\end{equation}
which indicates that although $\pmb{h}$ remains conserved in the flight phase, the average angular velocity $\pmb{\bar{\omega}}$ can be regulated by manipulating the actuated part $\prescript{\bar{\omega}}{}{\pmb{I}}_{h}^a(\pmb{q})$ of the \ac{ccrbi}, thereby influencing the overall body posture.

\subsubsection{MPC Formulation}
$\prescript{\bar{\omega}}{}{\pmb{I}}_{h}^{a}(\pmb{q})$ has a complex nonlinear relationship with the robot configuration $\pmb{q}$. For simplification, we consider the actuated leg as a constant density cuboid with variable dimension, resulting in a diagonal inertia matrix that can be estimated through the leg length $\pmb{\xi}$, defined as 
\begin{equation}
    \pmb{\xi}=\pmb{r}-\pmb{\rho}.
    \label{eq:leg_length}
\end{equation}
Furthermore, we define the Euler angle vector between the \ac{com} coordinate frame and the leg length vector $\pmb{\xi}$ as $\pmb{\theta}$, expressed in the X-Y-Z convention. 
Since the floating-base part maintains a constant inertia $\prescript{\bar{\omega}}{}{\pmb{I}}_{h}^{f}$ during the motion, we have:
\begin{equation}
    \prescript{\bar{\omega}}{}{\pmb{I}}_{h}=\prescript{\bar{\omega}}{}{\pmb{I}}_{h}^{f}+\prescript{\bar{\omega}}{}{\pmb{I}}_{h}^{a}\approx \prescript{\bar{\omega}}{}{\pmb{I}}_{h}^{f} + {\pmb{R}(\pmb{\theta})}^{\mathsf{T}}\,\bar{\pmb{I}}_{\xi}\,{\pmb{R}(\pmb{\theta})},
    \label{eq:inertia_approx}
\end{equation}
where
\begin{equation}
\label{eq:inertia_calibration}
\bar{\pmb{I}}_{\xi}=
\begin{bmatrix}
    k_{\xi1} & 0 & 0\\
    0 & k_{\xi2}& 0\\
    0 & 0 & k_{\xi3}\\
\end{bmatrix}
\begin{bmatrix}
    \left\| \xi \right\| ^2 & 0 & 0\\
    0 & \left\| \xi \right\| ^2& 0\\
    0 & 0 & 1\\
\end{bmatrix},
\end{equation}
and $\pmb{R}(\cdot)$ represents the transformation from Euler angles to a standard rotation matrix, $\pmb{k}_{\xi}=diag(k_{\xi1},k_{\xi2},k_{\xi3})$ is a diagonal matrix that can be calibrated with the multi-body model of actuated leg.

Define the integral of $\pmb{\bar{\omega}}$ with time as equimomental ellipsoid orientation $\pmb{\mathcal{L}}\in\mathbb{R}^{3\times1}$:
\begin{equation}
    \pmb{\mathcal{L}}_{\left[t_1\right]}- \pmb{\mathcal{L}} _{\left[t_2\right]}=\int_{t_1}^{t_2}{ \pmb{\bar{\omega}}}\,dt=\int_{t_1}^{t_2}{{(\prescript{\bar{\omega}}{}{ \pmb{I}}_{h}^{f}+\prescript{\bar{\omega}}{}{ \pmb{I}}_{h}^{a})}^{-1} \pmb{h}}\,dt.
\label{eq:lll}
\end{equation}
Discretize \cref{eq:lll} at $k^{th}$ time interval and utilize approximated inertia \cref{eq:inertia_approx}, we can have the following equality constraint:
\begin{equation}
\pmb{\mathcal{L}}_{\left[k\right]}=\sum_{i=1}^k{\left(\prescript{\bar{\omega}}{}{\pmb{I}}_{h}^{f} + \prescript{\xi}{}{\pmb{R}(\pmb{\xi})}^{\mathsf{T}}\,\bar{\pmb{I}}_{\xi}\,\prescript{\xi}{}{\pmb{R}(\pmb{\xi})}\right)^{-1}\pmb{h}_{\left[k\right]}}+\pmb{\mathcal{L}}_{\left[0\right]}.
    \label{eq:ellipsoidOri}
\end{equation}

Choosing the set of decision variables as:
\begin{equation}
\resizebox{0.89\linewidth}{!}{
$\begin{aligned}
    &\pmb{\Gamma}^{\text{mpc}}=\{\pmb{r}_{\left[k\right]},\pmb{\dot{r}}_{\left[k\right]}, \pmb{\ddot{r}}_{\left[k\right]}, \pmb{\mathcal{L}}_{\left[k\right]}, \pmb{h}_{\left[k\right]}, \pmb{\dot{h}}_{\left[k\right]}, \pmb{f}_{\rho {\left[k\right]}}^i, \pmb{\rho}_{\left[k\right]}^i,\beta_{\left[k\right]}^{ij}
,\\ &\forall i=1,\cdots,N_c, j=1,\cdots,N_d, k=1, \cdots,N \} \in\mathbb{R}^{58\times N},
\end{aligned}$}
\end{equation}
we formulate the real-time \ac{mpc} as follows: 
\begin{equation}  
\resizebox{0.89\linewidth}{!}{
$\begin{aligned}
   &\min_{\pmb{\Gamma}^{\text{mpc}}} \quad \sum_{k=1}^N{\left( \sum_{i=1}^{N_c}{\left\|\pmb{f}_{\rho\left[k\right]}^i\right\|^2}+\left\| \pmb{\ddot{r}}_{\left[k\right]}\right\|^2+\left\|\pmb{\dot{h}}_{\left[k\right]}\right\|^2\right)}+\left\| \pmb{\mathcal{L}}_{\left[N\right]}-\pmb{\mathcal{L}}_{\left[N\right]}^* \right\|^2\\
    &\textit{s.t.} \quad\quad\quad\quad\quad\quad   \text{\cref{eq:planning_con1,eq:planning_con2,eq:planning_confircton1,eq:planning_confircton2,eq:planning_con6,eq:planning_con7,eq:planning_con8,eq:ellipsoidOri}}
    \end{aligned}$}
\label{eq:landingmpc}
\end{equation}  
This formulation utilizes the feedback signal from the state estimator as the initial value. While similar to the trajectory optimization problem \cref{eq:trajopt}, it eliminates $\pmb{q}$, $\pmb{\dot{q}}$ in decision variables, along with the complex nonlinear constraints \cref{eq:planning_con3,eq:planning_con4,eq:planning_con5}. These modifications lead to a significantly faster convergence speed. $\pmb{\mathcal{L}}_{\left[0\right]}$ and $\pmb{\mathcal{L}}_{\left[N\right]}^*$ are the manually specified equimomental ellipsoid orientation at the first and last time intervals.

\subsection{Centroidal Momentum-based IK} 
To accurately track the optimized centroidal momentum and equimomental ellipsoid orientation obtained from the \ac{mpc}, we designed a centroidal momentum-based \ac{ik} method. A primary challenge in mapping the centroidal trajectory is the angular velocity of the
frame attached to the \ac{com} can not be integrated due to the non-constant nature of \ac{ccrbi} as introduced before~\cite{saccon2017centroidal}.

\subsubsection{Instantaneous Mapping Method}
To address this issue, our initial approach first maps the centroidal momentum to general velocity $\pmb{\dot{q}}$ based on the previous general position $\pmb{q}^{\text{pre}}$, and then integrates $\pmb{\dot{q}}$ to obtain new general position $\pmb{q}$, resulting in an optimization problem in each control cycle:
\begin{equation}
\begin{aligned}
    &\underset{\pmb{q},\pmb{\dot{q}}}{\min}\,\,\pmb{W}_{cm}\left\| \left. 
        \pmb{H}^{\text{mpc}} - \pmb{A}(\pmb{q})\pmb{\dot{q}} \right\| \right.+\pmb{W}_{\dot{q}}\left\| \left. \pmb{\dot{q}} \right\| +\pmb{W}_q\left\| \left. \pmb{q}-\pmb{q}^{*} \right\| \right. \right. \\
     &\textit{s.t.}\quad\quad\quad\quad\quad\quad\quad    \pmb{q}=\pmb{q}^{\text{pre}}+\pmb{\dot{q}}\varDelta t\\
    &\quad\quad\quad\quad\quad\quad\quad\quad \pmb{r}^{\text{mpc}}=\pmb{\mathcal{F}}_r(\pmb{q}) \\
    &\quad\quad\quad\quad\quad\quad\quad\quad \pmb{\rho}^{\text{mpc}} = \pmb{\mathcal{F}}_\rho(\pmb{q}) \\
\end{aligned}
\label{eq:instances}
\end{equation}
where the superscript ${[\cdot]}^\text{mpc}$ is the optimized value of a variable from the \ac{mpc}, $\pmb{H}$ is defined in \cref{eq:cmcal}, $\pmb{q}^{*}$ is the nominal posture of the robot. $\pmb{W}_{cm}$, $\pmb{W}_{\dot{q}}$, and $\pmb{W}_q$ are weighting matrices.

\subsubsection{Simplified Mapping Method} 
However, simultaneously solving for $\pmb{q}$ and $\pmb{\dot{q}}$ leads to slow computational speeds. Therefore, we simplify the problem by choosing $\pmb{\dot{q}}$ as the only decision variable and solve $\pmb{q}$ through integration.
\begin{equation}
\begin{aligned}
    &\underset{\pmb{\dot{q}}}{\min}\quad\pmb{W}_{cm}\left\| \left. \pmb{H}^{\text{mpc}} +\pmb{H}^{\text{aug}}- \pmb{A}(\pmb{q})\pmb{\dot{q}} \right\| \right. +\pmb{W}_{\dot{q}}\left\| \left. \pmb{\dot{q}} \right\| \right. \\
    &{\text{s.t.}\quad\quad\quad\quad\quad\quad \pmb{J}\pmb{\dot{q}}=\pmb{\dot{\rho}}^{\text{des}} + \pmb{\dot{\rho}}^{\text{aug}}}\\
\end{aligned}
\label{eq:cmik}
\end{equation}
The cost function of this simplified optimization problem accounts for the centroidal momentum and generalized velocity while the constraint limits the foot endpoint velocity. $\pmb{\dot{\rho}}^{\text{des}}$ is the desired foot endpoint speed acquired from the landing heuristic controller (to be introduced next), 
\begin{equation}
    \pmb{\dot{\rho}}^{\text{aug}}=\pmb{k}_{\rho}\left( \pmb{\rho}^{\text{mpc}}-\pmb{\rho}^{\text{fk}} \right)
\end{equation} 
is a feedback augment to improve the foot endpoint trajectory tracking where the superscript ${[\cdot]}^\text{fk}$ denotes a variable calculated through forward kinematics. 
\begin{equation}
\resizebox{0.89\linewidth}{!}{
$\pmb{H}^{\text{aug}} = 
[\pmb{k}_{r}(\pmb{r}^{\text{mpc}\mathsf{T}} - \pmb{r}^{\text{fk}\mathsf{T}}),\pmb{k}_{R}( \pmb{R}^{\text{fk}\mathsf{T}} \pmb{R}^{*}-\pmb{R}^{*\mathsf{T}}\pmb{R}^{\text{fk}})_\vee]^\mathsf{T}$}
\end{equation} 
is a feedback augment to improve the \ac{com} and torso orientation trajectory tracking, where $\pmb{R}^{*}$ stands for the nominal torso orientation expressed as standard rotation matrix, 
and $\left(\cdot\right)_\vee$ is the mapping from SO(3) to $\mathbb{R}^3$, $\pmb{k}_{\rho}$, $\pmb{k}_{r}$, $\pmb{k}_{R}$ are the feedback gains.

\subsection{Landing Heuristic Controller}

\begin{figure*}[b!]
    \centering
    \begin{subfigure}[b]{0.4\linewidth}
    \centering
    \includegraphics[width=\linewidth]{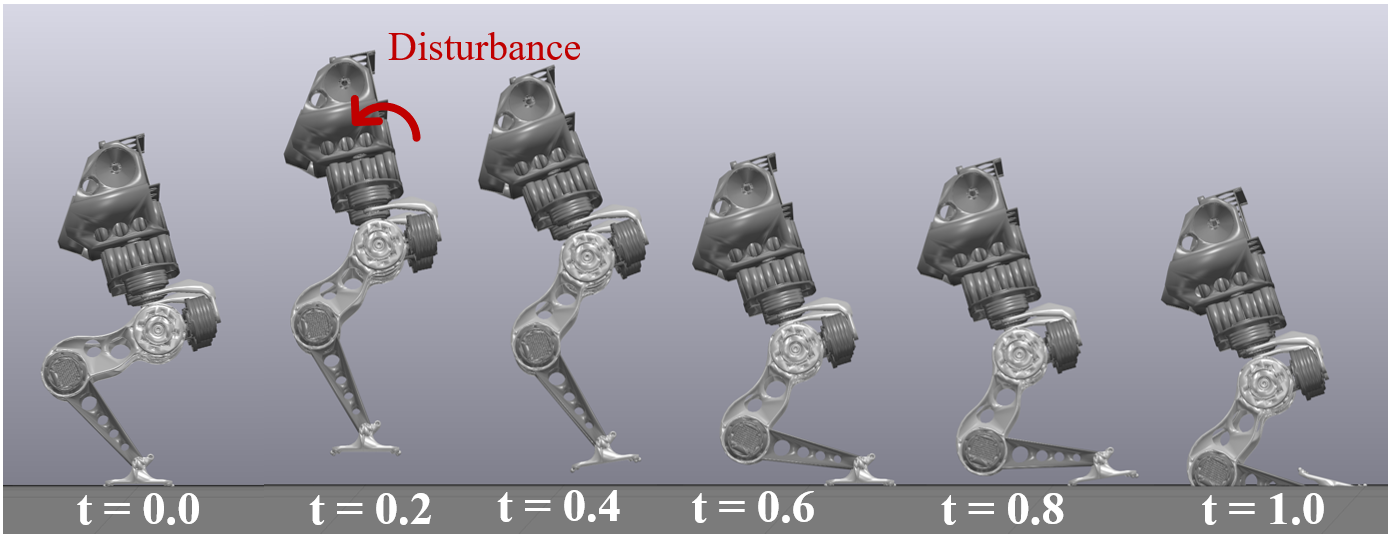}
    \caption{SRBM-MPC}
    \label{fig:SRBM-MPC}
    \end{subfigure}
    \begin{subfigure}[b]{0.28\linewidth}
    \centering
        \includegraphics[width=\linewidth,trim=0cm 0cm 1.8cm 1.5cm,clip]{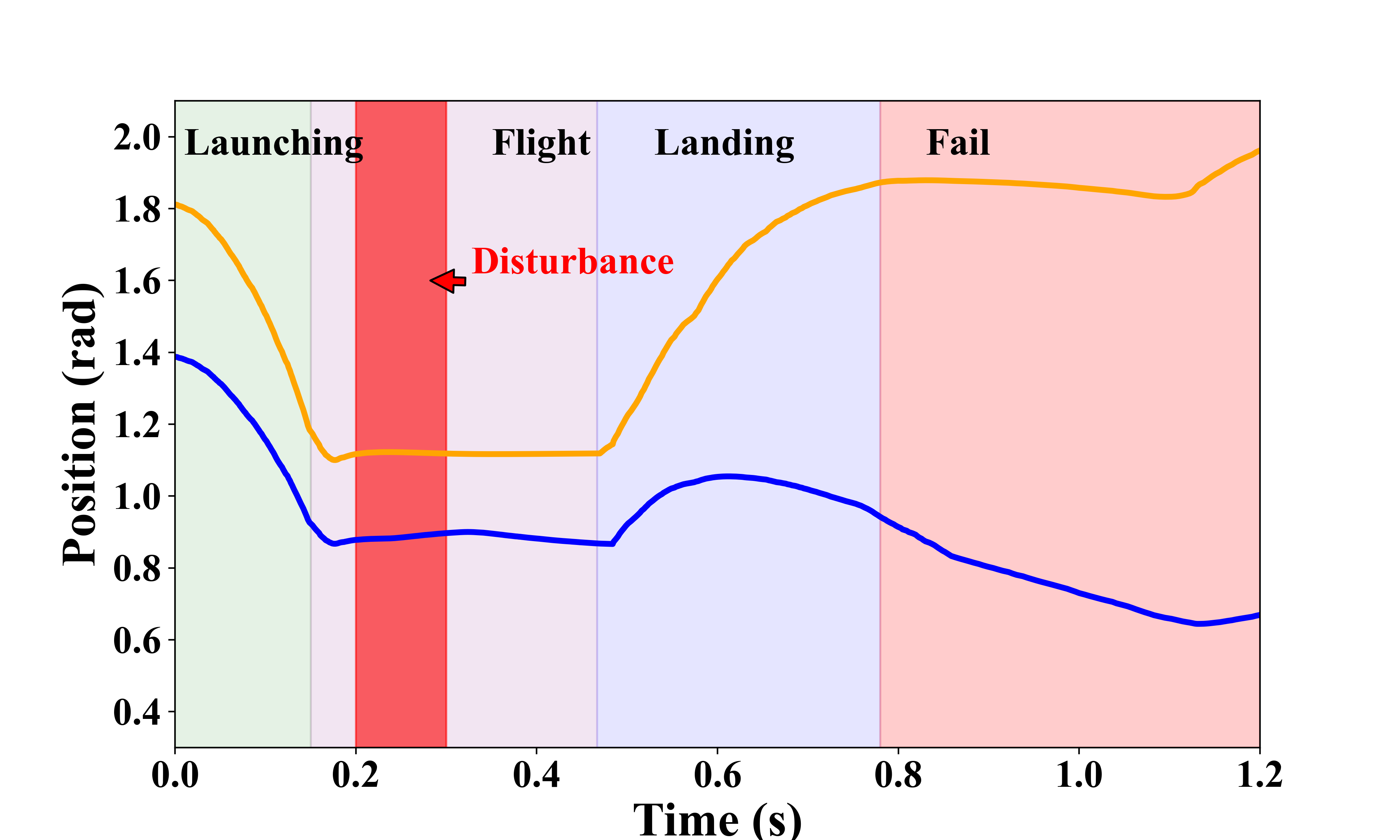}
        \caption{SRBM-MPC leg configuration}
    \label{fig:SRBM-MPC_leg}
    \end{subfigure}
    \begin{subfigure}[b]{0.3\linewidth}
    \centering
        \includegraphics[width=\linewidth,trim=0cm 0.2cm 0cm 0.2cm,clip]{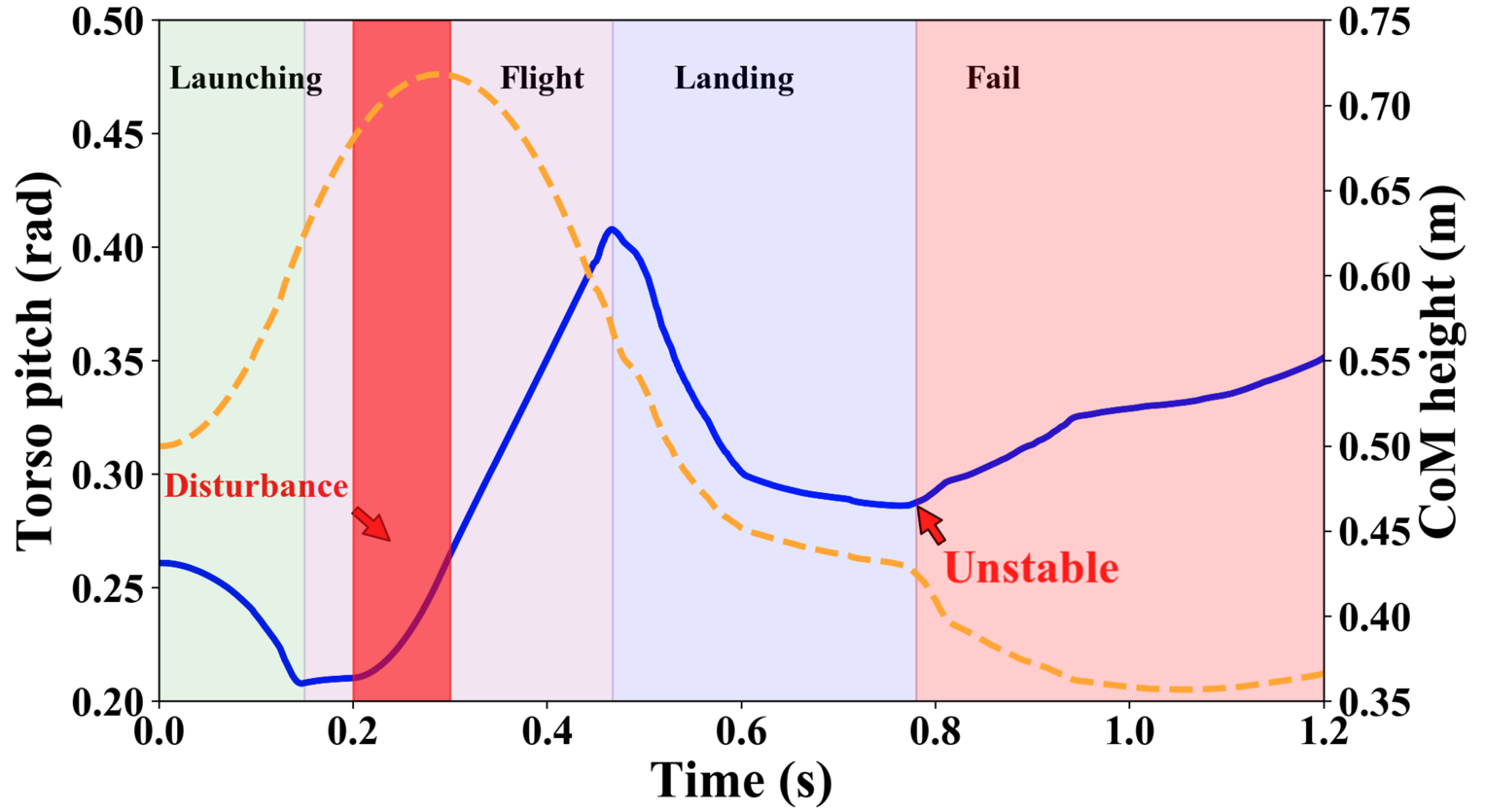}
        \caption{SRBM-MPC jumping data}
    \label{fig:SRBM-MPC_curve}
    \end{subfigure}\\
    \begin{subfigure}[b]{0.4\linewidth}
    \centering
        \includegraphics[width=\linewidth]{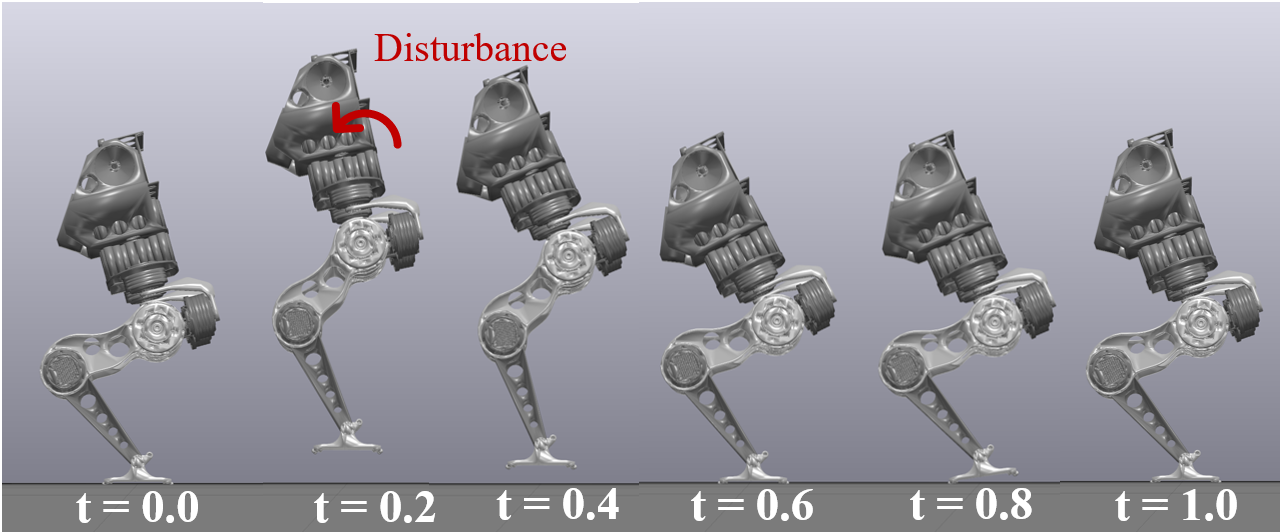}
        \caption{CDM-MPC}
    \label{fig:CDM-MPC}
    \end{subfigure}
    \begin{subfigure}[b]{0.28\linewidth}
    \centering
        \includegraphics[width=\linewidth,trim=0cm 0cm 1.8cm 1.5cm,clip]{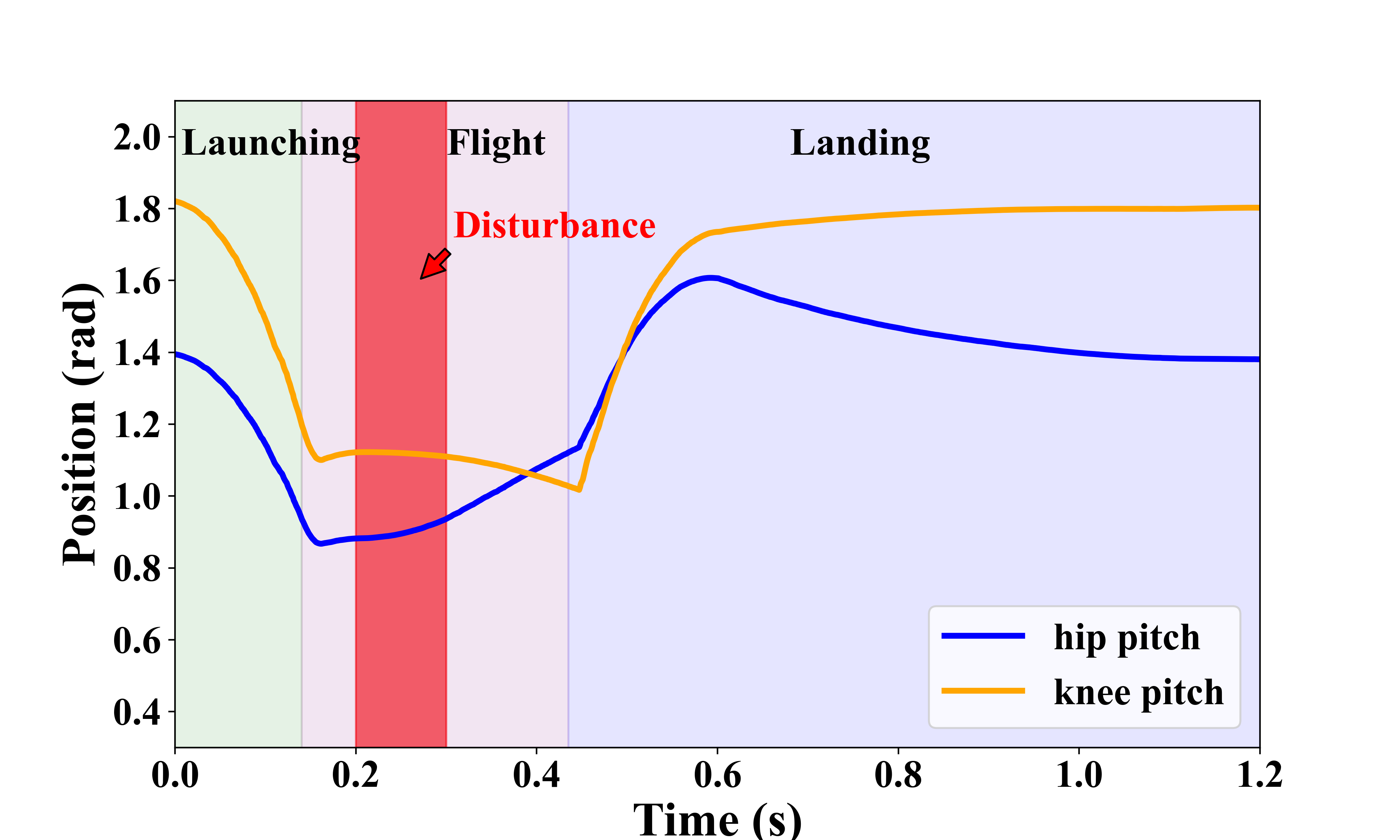}
        \caption{CDM-MPC leg configuration}
    \label{fig:CDM-MPC_leg}
    \end{subfigure}
    \begin{subfigure}[b]{0.3\linewidth}
    \centering
        \includegraphics[width=\linewidth,trim=0cm 0.2cm 0cm 0.2cm,clip]{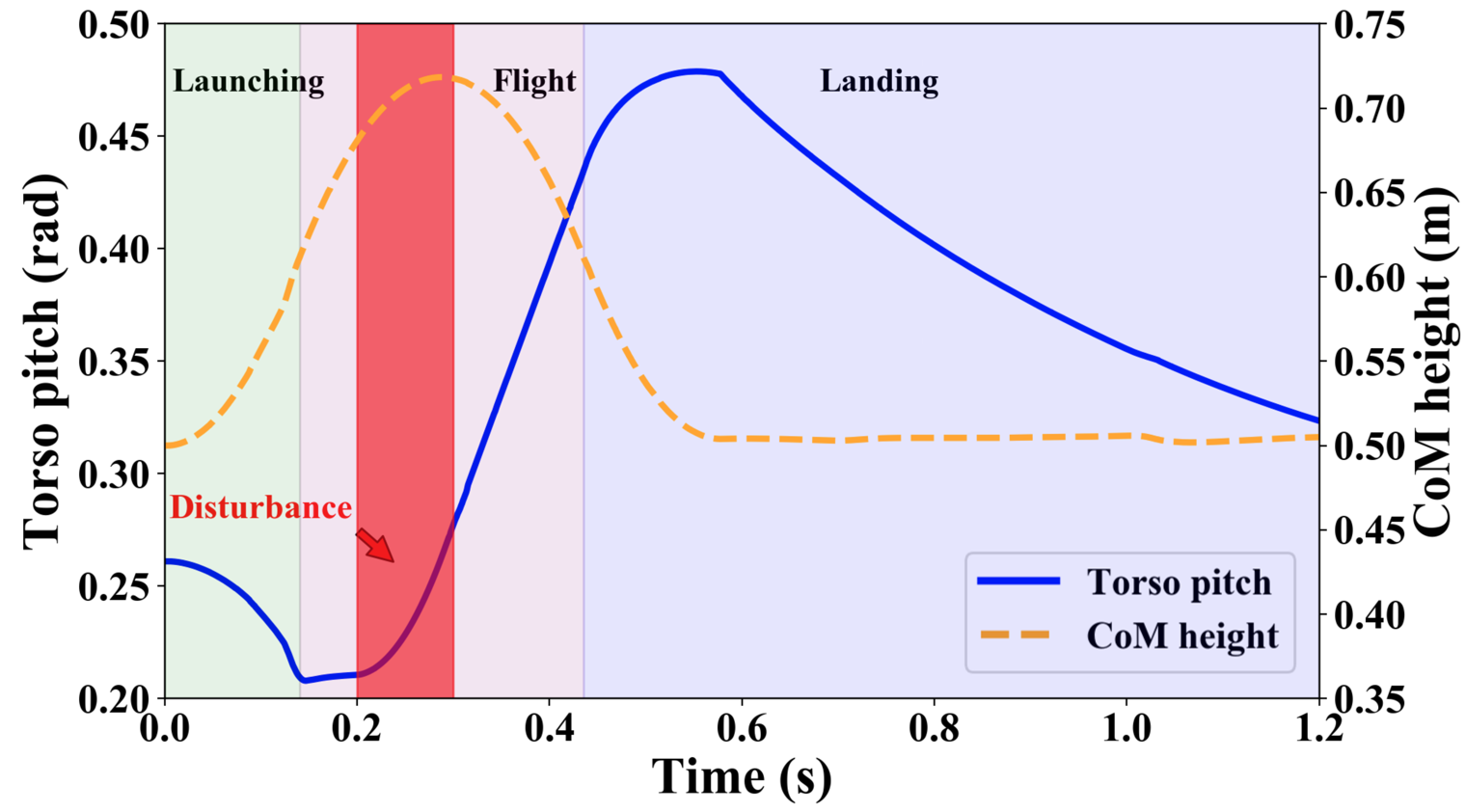}
        \caption{CDM-MPC jumping data}
    \label{fig:CDM-MPC_curve}
    \end{subfigure}
    \caption{\textbf{Case1 (Simulation): Disturbance rejection performance study during in-place jumping.} (a) The SRBM-MPC method cannot maintain the robot's stability when subjected to disturbance torque, leading to its collapse. (c) Conversely, the CDM-MPC method preserves the robot's stability throughout the entire flight phase, culminating in a successful landing. (b)(e) depict the joint configuration of hip pitch and knee pitch joints while (c)(f) depict the pitch angle of the torso and the CoM height for both methods, respectively.}
    \label{fig:case1}
\end{figure*}

The landing angle $\pmb{\theta}$ and leg length $\pmb{\xi}$ are critical for landing stability, and thus are controlled by the PD feedback controllers. As the CoM is uncontrollable in the air, we control the foot endpoint position to achieve the desired landing angle and leg length, the control law is designed as follows: 
\begin{equation}
\resizebox{0.83\linewidth}{!}{
$\begin{aligned}
\pmb{\xi}^{\text{des}}&=\pmb{\xi }^{\text{mpc}}+\pmb{k}_{p}^{\xi}\left( \pmb{\xi}^{\text{mpc}}-\pmb{\xi }^{\text{est}} \right) +\pmb{k}_{d}^{\xi}\left( \pmb{\dot{\xi}}^{\text{mpc}}-\pmb{\dot{\xi}}^{\text{est}} \right) 
\\
\pmb{\theta}^{\text{des}}&=\pmb{\theta}^{\text{mpc}}+\pmb{k}_{p}^{\theta}\left( \pmb{\theta}^{\text{mpc}}-\pmb{\theta}^{\text{est}} \right) +\pmb{k}_{d}^{\theta}\left( \pmb{\dot{\theta}}^{\text{mpc}}-\pmb{\dot{\theta}}^{\text{est}} \right) 
\\
\pmb{\rho}^{\text{des}}&=\pmb{r}^{\text{est}}+\left\| \pmb{\xi}^{\text{des}}\right\|/\left\| \pmb{\theta}^{\text{des}}\right\| \pmb{\theta}^{\text{des}}
\end{aligned}$}
\label{eq:rho_feedback}
\end{equation}
where the superscript ${[\cdot]}^\text{est}$ is the estimated value of a variable from the state estimator, 
the optimized leg length $\pmb{\xi}^{\text{mpc}}$ is calculated with MPC decision variable set $\pmb{\Gamma}^{\text{mpc}}$ and \cref{eq:leg_length}, which is then utilized to calculate the optimized 
landing angle $\pmb{\theta}^{\text{mpc}}$, $\pmb{k}_{p}^{\xi}$,  $\pmb{k}_{d}^{\xi}$, $\pmb{k}_{p}^{\theta}$,  $\pmb{k}_{d}^{\theta}$ are feedback gains.


\subsection{Whole-body Controller} 
Our study employs a typical weighted \ac{wbc} controller to regulate the movements of the entire robot~\cite{feng2014optimization}. Distinct weights are set for tracking the trunk and foot trajectories of the robot. The robot's reference positions and velocity trajectories are obtained through the centroidal momentum-based IK, while the reference accelerations are derived by differentiating velocities. Additionally, we integrate a constraint in line with the law of conservation of momentum~\cite{vatavuk2021precise}, aiming to achieve improved tracking of the centroidal angular momentum. This integration enhances the robot's dynamic balance and stability during complex maneuvers.

\begin{table}[b!]
\small
\centering
\caption{\textbf{Planing \& Control Parameters}}
\resizebox{0.83\linewidth}{!}{
\begin{tabular}{ccc}
\toprule
\textbf{Group} & \textbf{Parameters} & \textbf{Value}\\
\midrule
\multirow{5}{*}{\rotatebox[origin=c]{0}{\makecell{\textbf{KMP}~\cref{eq:trajopt}}}}
& Step time & $50~ms$\\
& Horizon $N$ & $2~s$\\
& $f_{\text{max}}$ & $2000 N$\\
& $\xi_{\text{min}}$ & $0.4 m$\\
& Solve time & $1\sim10~s$ \\
\midrule
\multirow{6}{*}{\rotatebox[origin=c]{0}{\makecell{\textbf{Real-time MPC}~\cref{eq:landingmpc}}}}
& Step time & $10~ms$\\
& Horizon $N$ & $1~s$\\
&  $f_{\text{max}}$ & $2000 N$\\
&  $\xi_{\text{min}}$ & $0.4 m$\\
& $k_{\xi1},k_{\xi2},k_{\xi3}$ & $0.51,0.62,0.07$ \\
& Solve time & $\approx20~ms$\\
\midrule
\textbf{CMIK}~\cref{eq:instances} & Solve time & $\approx2.5~ms$ \\
\multirow{2}{*}{\rotatebox[origin=c]{0}{\makecell{\textbf{CMIK}~\cref{eq:cmik}}}}
& Solve time & $\approx1~ms$\\
& $\pmb{k}_{\rho},\pmb{k}_{r},\pmb{k}_{R}$ &  $[10.0,10.0,1.0]*\pmb{1}_3$\\
\midrule
\multirow{3}{*}{\rotatebox[origin=c]{0}{\makecell{\textbf{Landing}~\cref{eq:rho_feedback}}}}
& $\pmb{k}_{p}^{\xi},\pmb{k}_{d}^{\xi}$ & $[0.1,0.03]*\pmb{1}_3$\\
& $\pmb{k}_{p}^{\theta},\pmb{k}_{d}^{\theta}$ & $[0.5,0.2]*\pmb{1}_3$\\
& Solve time & $\leq0.1~ms$\\
\bottomrule
\end{tabular} }
\label{tab:para}
\end{table}

\section{Verification} \label{sec:verification}
To rigorously evaluate the efficacy of the proposed CDM-MPC framework, simulations and experiments were executed using the KUAVO bipedal robot platform. The dynamic simulator was developed in the Drake simulation environment~\cite{drake} and utilized the KUAVO's detailed multibody model. Besides incorporating the mass and inertia parameters reflecting the physical robot's specifications, the simulator also accounted for the torque saturation and motion range of each motor, control frequency, and measurement noise, as described in~\cref{sec:hardware_introduction}. Physical validation trials were carried out on the physical KUAVO platform. Main control parameters are summarized in \cref{tab:para}.

Four test cases were designed as follows: (i) \textbf{Case 1:} we compared the disturbance rejection performance of the proposed CDM-MPC method with a baseline SRBM-MPC method during in-place jumping; (ii) \textbf{Case 2:} we studied the robustness of proposed landing controller with variable forward jumping velocities; (iii) \textbf{Case 3:} we validated the proposed framework on the physical KUAVO robot in a jumping experiment; and (iv) \textbf{Case 4:} we explored the versatility of the framework by applying it to walking locomotion.  

\subsection{Inertia Parameter Calibration}

\begin{figure}[t!]
  \centering
  \includegraphics[width=\linewidth, trim=0.3cm 0.2cm 0cm 0.2cm, clip]{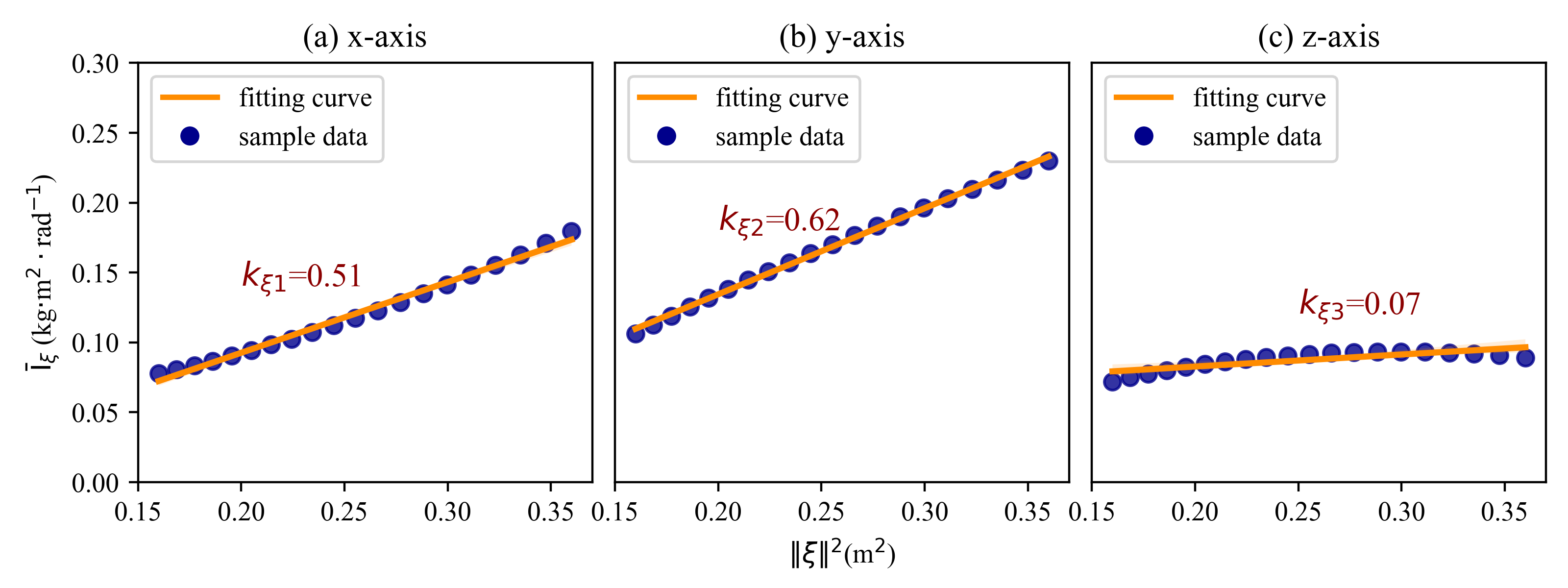}
  \caption{\textbf{Inertia parameter calibration results.} The $x$ and $y$ axes components of $\bar{\pmb{I}}_{\xi}$ has linear relationship with $\left\| \xi \right\|^2$, while the $z$ axis component of $\bar{\pmb{I}}_{\xi}$ is invariant.}
  \label{fig:calibration}
\end{figure}

To calibrate the inertia parameter matrix $\pmb{k}_{\xi}$ in \cref{eq:inertia_calibration}, 20 robot configuration data points were collected in the simulator with different leg lengths from $0.4$ to $0.6~m$. Utilizing the calculation of actuated part $\prescript{\bar{\omega}}{}{\pmb{I}}_{h}^a(\pmb{q})$ of the \ac{ccrbi} in \cref{eq:inertia_decomposition} as ground truth, \cref{eq:inertia_calibration,eq:inertia_approx} are utilized to calculate the calibration matrix $\pmb{k}_{\xi}$. The calibration results are illustrated in \cref{fig:calibration}, where the component of $\bar{\pmb{I}}_{\xi}$ in $x$ and $y$ axes have an obvious linear relationship with $\left\| \xi \right\|^2$, while the component in $z$ axis is almost invariant to the changes of leg length. Subsequently, the parameter matrix is calculated from the sampled data by leveraging a linear regression method. The calibrated parameters are included in \cref{tab:para}.

\begin{figure}[ht!]
    \centering
    \includegraphics[width=\linewidth,trim=1cm 0cm 0.5cm 0.5cm, clip]{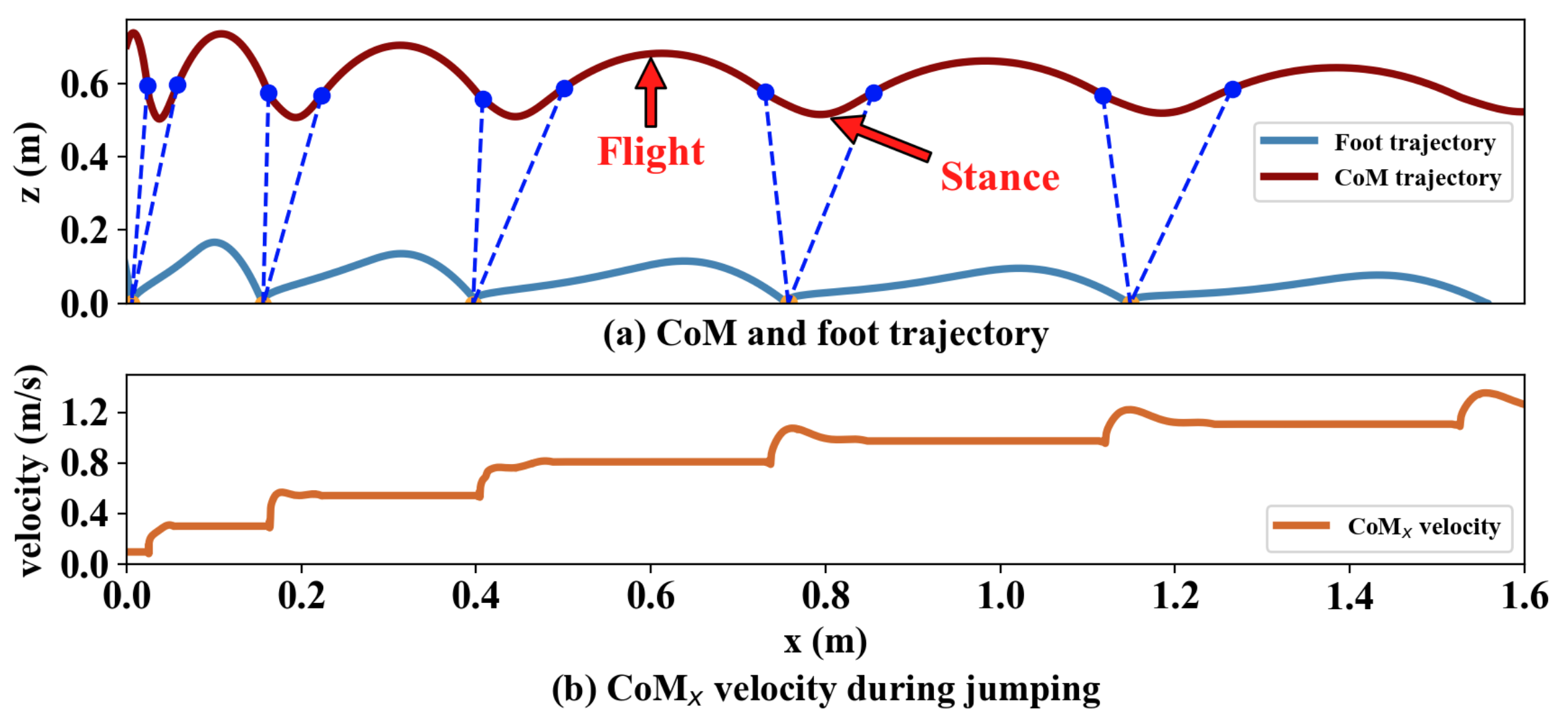}
    \caption{\textbf{Case2 (Simulation): Continuous forward jumping with increasing jumping velocities.} (a) CoM and foot trajectory of the robot during the process. (b) Corresponding estimated velocity.}
    \label{fig:case2}
\end{figure}
\subsection{In-Place Jumping Simulation}
To verify the effectiveness of the CDM-MPC method in rejecting disturbances, we compared it with a baseline SRBM-MPC method. The simulation results are shown in \cref{fig:case1}. Both methods employed the same optimized in-place jumping trajectory from the kinodynamic motion planning. At the apex of the robot's flying phase, a perturbing torque with a magnitude of $20~Nm$ with a duration of $0.1~s$ was applied at the robot's torso as an unknown disturbance.

As shown in \cref{fig:SRBM-MPC}, with the SRBM-MPC method, the body of the robot gradually tilts after the injected disturbance during the flight phase, resulting in an unstable landing. In contrast, as shown in \cref{fig:CDM-MPC}, the angular momentum-based control framework, which takes into account leg inertia changes, facilitates refined control over the robot's posture. During the flight phase, both body and leg postures are dynamically adjusted in accordance with a trajectory generated by the CDM-MPC and the momentum-based IK. This approach improves the in-flight disturbance-rejection capability of the robot, ensuring a stable landing.

\begin{figure}[t!]   
    \centering
    \begin{subfigure}[b]{0.9\linewidth}
    \centering
        \includegraphics[width=\linewidth]{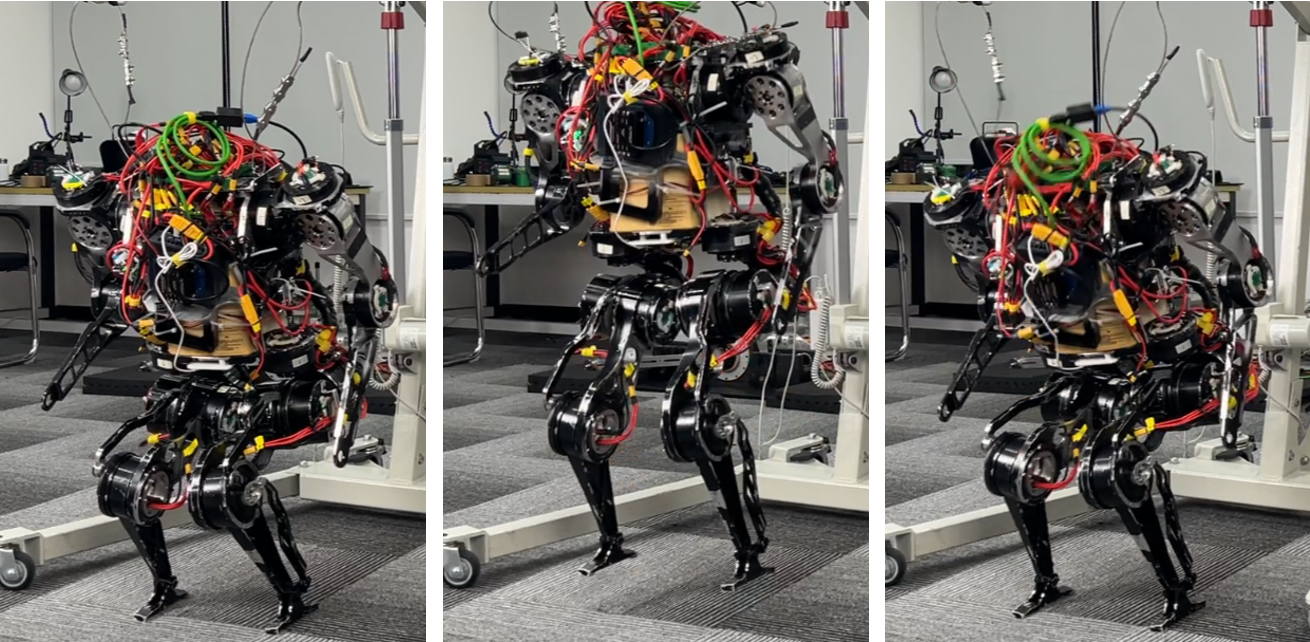}
        \caption{Key frames in experiment.}
    \label{fig:exp_clips}
    \end{subfigure} \\
    \begin{subfigure}[b]{0.9\linewidth}
    \centering
    \includegraphics[width=\linewidth]{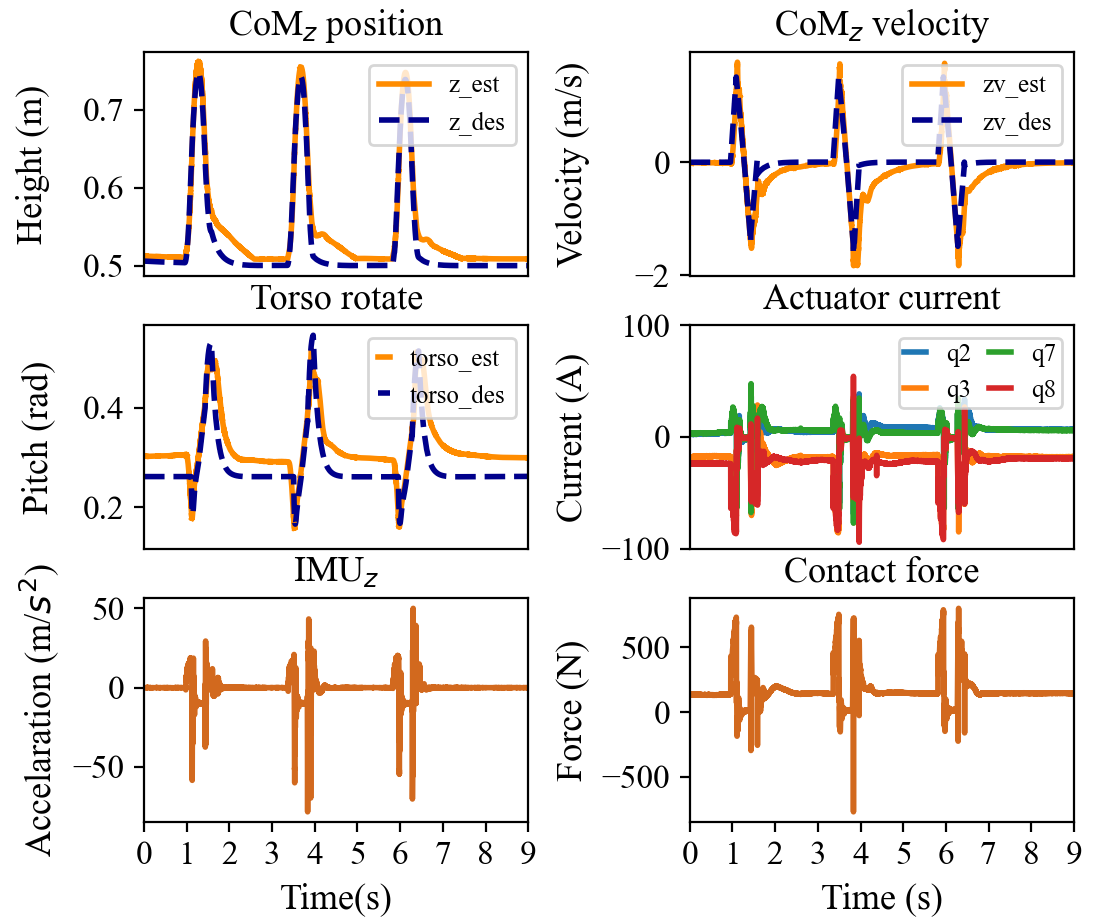}
    \caption{Experiment results}
    \label{fig:experiment_curve}
    \end{subfigure}
    \caption{\textbf{Case3 (Experiment): Continuous in-place jumping on physical KUAVO robot.} The KUAVO robot executes three vertical jumps of approximately $0.28~m$ and lands stably with the proposed jump control framework.}
    \label{fig:case3}
\end{figure}

\subsection{Forward Jumping Simulation}
In the forward jumping simulation, a range of initial jump trajectories corresponding to velocities of $0.5,1.0,1.5,2.0~m/s$ were provided to serve as starting points for optimization. The CDM-MPC algorithm is designed to automatically select the trajectory that aligns most closely with the commanded velocity as its initial guess, thereby improving the efficiency of the trajectory planning process. The robot's key performance metrics and jumping velocity commands throughout the jumping sequence are illustrated in \cref{fig:case2}. The corresponding visualization is shown in \cref{fig:motiv}.

The results demonstrated (i) the robustness of the landing controller which makes subtle adjustments to regulate the landing angle as the robot descends (ii) the capability of the CDM-MPC framework to adaptively generate a range of jump trajectories from a pool of precomputed offline samples, facilitating smooth transitions between various jumping velocities.

\subsection{Physical Jumping Experiment}

In the experiment, we validate our proposed jumping framework using KUAVO, our full-sized humanoid robot platform, as detailed in \cref{sec:dynamics}. As shown in \cref{fig:experiment_curve}, the IMU-measured acceleration curve indicates significant impacts on the robot's torso during both the launching and landing phases. The contact forces are derived directly from joint current readings, which are significantly affected by joint current noise at the moment of landing, resulting in notable spikes. Despite these challenges, the robot's stable landing performance showcases our control method's robustness to impacts and noise, highlighting our framework's effectiveness in physical humanoid robot jumping dynamics. The keyframes are shown in \cref{fig:exp_clips}.

\subsection{Walking Experiment}

\begin{figure}[t!]
    \centering
    \includegraphics[width=\linewidth,trim=0cm 0cm 0cm 0cm, clip]{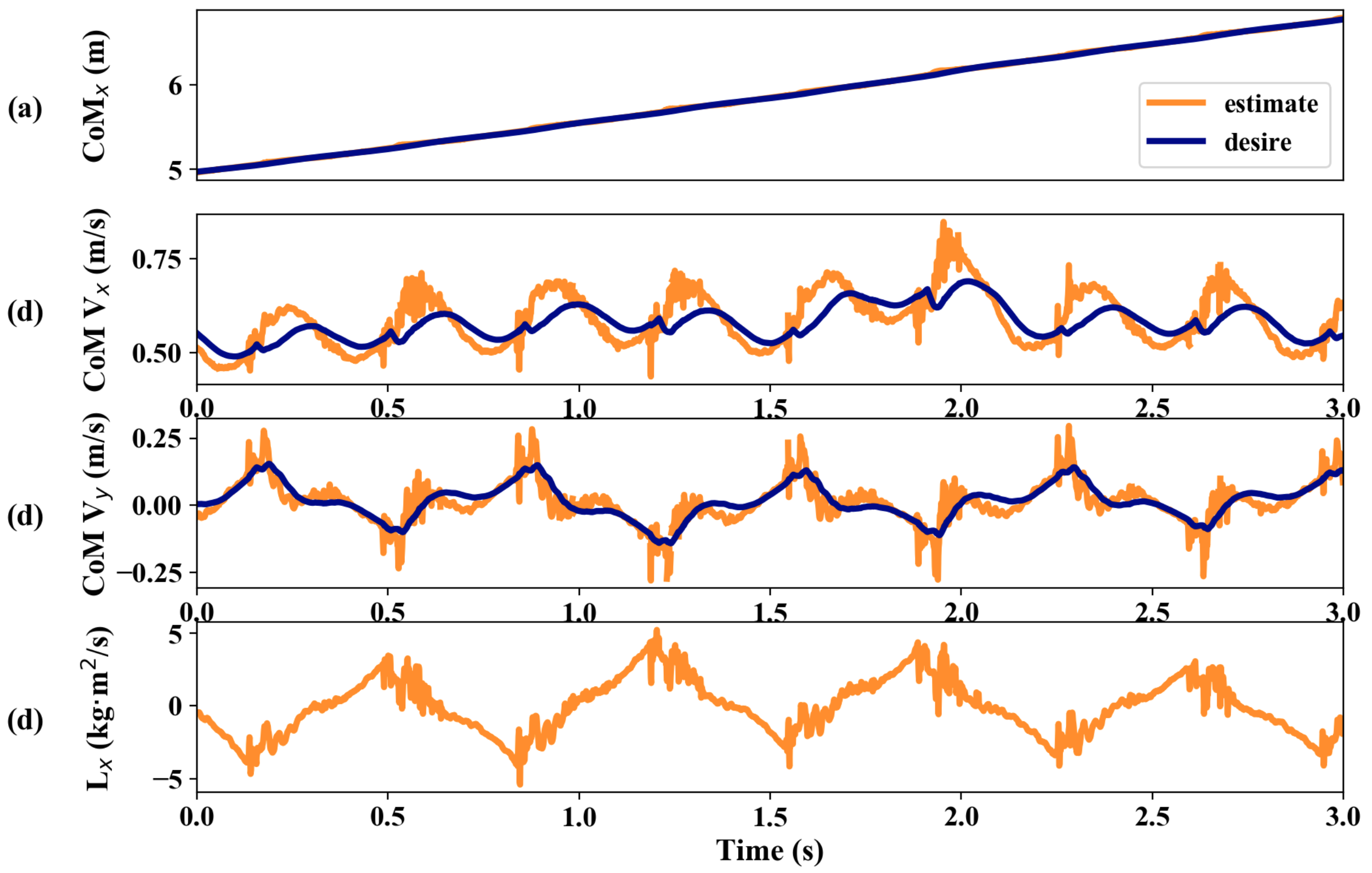}
    \caption{\textbf{Case4 (Experiment): Apply the proposed framework for walking control.} The KUAVO robot achieves stable walking performance with a speed of $0.6~m/s$.}
    \label{fig:case4}
\end{figure}

As shown in \cref{fig:case4}, to demonstrate the generalizability of our framework, we applied it to walking control. The robot walks stably at $1.2~m/s$ in simulation and $0.6~m/s$ on hardware.  This verifies that the framework can be extended to other locomotion modes.

\section{Conclusion} \label{sec:conclusion}
In this paper, we have presented an integrated dynamic planning and control framework (CDM-MPC) for the jumping motion of bipedal robots. This framework considers centroidal momentum in both dynamics planning of the launching phase and online tracking control of the flight phase, and it integrates a robust landing to ensure stability. Altogether, the proposed framework enables agile and continuous jumping motions on full-sized bipedal robots. We validated the effectiveness of this framework based on a novel full-sized bipedal robot KUAVO in both realistic simulations and real-world experiments and confirmed its applicability to other locomotion modes, such as walking. Future directions include generalizing the framework for unified walking, running, and jumping control with smooth transitional behavior and integrating reinforcement learning-based methods for performance and robustness improvement.

{
\small
\setstretch{0.96}
\bibliographystyle{ieeetr}
\bibliography{reference}
}
\end{document}